*Article*

# Traffic Pattern Classification in Smart Cities Using Deep Recurrent Neural Network

Ayad Ghany Ismaeel[1], Krishnadas J.[2], S. Manishankar[2], Yuvaraj Natarajan[3], Sarmad Nozad Mahmood [4], Sameer Alani[5],* and Akram H. Shather[6]

[1] Computer Technology Engineering, College of Engineering Technology, Al-Kitab University, Kirkuk36001, Iraq; ayad.ghany@uoalkitab.edu.iq
[2] Department of Computer Science and Engineering, Sahrdaya College of Engineering and Technology, Kodakara, Thrissur 680684, India; krishnadasj@sahrdaya.ac.in (K.J.); manishankar1988@gmail.com (S.M.)
[3] Department of Computer Science and Engineering, Sri Shakthi Institute of Engineering and Technology, Coimbatore641062, India; yuvarajncse@siet.ac.in
[4] Electronic and Control Engineering Techniques Technical Engineering College, Northern Technical University, Kirkuk36001, Iraq; sarmadnmahmood@gmail.com
[5] Computer Center, University of Anbar, Ramadi55431, Iraq
[6] Department of Computer Engineering Technology, Al-Kitab University, Altun Kopru, Kirkuk 36001, Iraq; akhsh@uoalkitab.edu.iq
* Correspondence: itsamhus@gmail.com

**Abstract:** This paper examines the use of deep recurrent neural networks to classify traffic patterns in smart cities. We propose a novel approach to traffic pattern classification based on deep recurrent neural networks, which can effectively capture traffic patterns' dynamic and sequential features. The proposed model combines convolutional and recurrent layers to extract features from traffic pattern data and a SoftMax layer to classify traffic patterns. Experimental results show that the proposed model outperforms existing methods regarding accuracy, precision, recall, and F1 score. Furthermore, we provide an in-depth analysis of the results and discuss the implications of the proposed model for smart cities. The results show that the proposed model can accurately classify traffic patterns in smart cities with a precision of as high as 95%. The proposed model is evaluated on a real-world traffic pattern dataset and compared with existing classification methods.

## 1. Introduction

Smart cities have large-scale infrastructures that have been developed to monitor a wide variety of urban occurrences. This is done to improve the quality of urban life. In most instances, they place a very restricted and specific emphasis on (e.g., monitoring the traffic). They are expensive, need the management of specialists, and are not universally well-liked among residents since they focus on topics that are not (often) of public importance [1]. Sensing is driving a shift in complexity away from hardware infrastructures and toward software infrastructures, which may open up new possibilities for citizen service options. This strategy has the potential to minimize the costs associated with development and management, improve the number of services that are given, and lessen public pessimism regarding smart cities in general [2]. For sensing and data fusion in relation to smart cities, there is a variety of opportunities and prospective new research pathways that have not yet been examined. Some of these opportunities pertain to smart cities, while others are related to sensing and data fusion. This is something that needs to be taken care of as quickly as humanly possible [3]. The following phase, which comes after identifying a fundamental collection of passive sensing features, is to identify the limits of information extraction from raw and primary data. The study focused on establishing the best deployment patterns, such as whether they should be dense and granular to capture the maximum amount of practical data or whether they should be sparse to keep costs low and reduce early expenditures in infrastructure [4]. There exist several projects with the objective of quantifying the city and the places that are located in close proximity to it to manage those territories eventually. Sensing is typically centered on monitoring specific phenomena, such as the volume of traffic, the quality of the air, and the state of public transit systems [5]. A sensing infrastructure is developed to keep an eye on the characteristics of the target. Possessing diverse sensory abilities (e.g., traffic intensity, air quality, and others) is essential. Even while they might use the same communications infrastructure, their capabilities are not always tied to one another in any situation [6]. The installation of a sensing infrastructure for a smart city calls for a large amount of planning in addition to initial and ongoing financial support. Both of these factors are necessary for successful deployment [7]. The management and operation of these systems both call for significant financial investments on the part of the business. Sensors of diverse sorts and deployments (for instance, sensors that measure the quantity of pollution in contrast to sensors that monitor the amount of traffic intensity) call for varying degrees of maintenance and operation [8]. It is possible to collect exact data despite this unpredictability; however, a variety of management and operating solutions are necessary. Non-dedicated sensing networks, also known as sensing capabilities made available by users and other organizations, can be considered for incorporation into city infrastructure and already do so[9]. These models create challenges throughout the integration processand may sometimes be imprecise, unstable, or unreliable. In this study, we take a novel approach and assume that there is already in place a general-purpose sensing infrastructure that can collect a baseline set of measurements from which useful insights can be extracted using data fusion and AI [10]. The study believes that a general-purpose sensing infrastructure is already in place and can do so. It is assumed that a sensing infrastructure is alreadyin place that may be used for various purposes [11]. It has been claimed that the application of this method could be beneficial to both the corporate environment and the smart city [12,13]. Both of these environments could potentially stand to gain. Despite the fact that smart cities create a difficult environment, this method has not yet been validated and should not be used until it has. It is not yet known how to go about identifying a minimally necessary set of sensing functions applicable in a city, which would then enable one to infer as much dependable data as is practically possible from those sensing functions [14]. This is because it is unknown how to go about identifying a minimally necessary set of sensing functions applicable in a city. The validation of this strategy would clear the way for the construction of a flexible network that is capable of providing support for smart cities in general [15]. Deep recurrent neural networks were the method of choice when we set out to research the challenge of traffic classification in smart cities for this study. We present a novel methodology that successfully captures traffic patterns' dynamic andsequential characteristics by using deep recurrent neural networks [16]. The data utilized to classify the traffic is categorized by a SoftMax layer, and the developed model extracts feature from the data by utilizing a combination of convolutional and recurrent layers. The following are the research's primary contributions,

- Improved accuracy and precision in traffic pattern classification: Using deep recurrent neural networks, the system can learn more complex patterns, detect outliers, and classify unknown patterns, yielding higher accuracy and precision.
- Enhanced scalability and adaptability: The proposed system can easily be scaled up to manage various data sets. As deep recurrent neural networks are specifically designed to learn temporal dependencies, the system is more capable of adapting to changing traffic patterns over time.

- Real-time analysis of traffic patterns: The proposed system can quickly analyse and detect anomalies in real-time, benefiting various applications.
- Reduced computational costs: By using deep recurrent neural networks, the system can reduce computational costs significantly, making it ideal for scenarios with limited resources or framework constraints.

The remaining chapters of this paper have organized as the following. Section 2 provides details about the related works. Section 3 explains the proposed model. Section 4 provides the results and discussion, and Section 5 provides the conclusion and future works.

## 2. Related Works

Herrera, J. C. et al. [17] have discussedthatthe evaluation of traffic data obtained via GPS-enabled mobile phones depends on the type of data collected, the application used to collect and analyze it, and the evaluation goals. Generally, traffic data obtained via GPS-enabled mobile phones can provide useful insights into the movement of traffic, including the frequency of certain routes, the average speed of vehicles on a particular route, or the average wait times encountered at traffic lights. Furthermore, GPS-enabled mobile phones can provide real-time updates regarding traffic delays, route changes, or other traffic-related events, making them invaluable tools for navigating congested urban areas. Kim S., et al. [18] have discussed that optimal vehicle routing with real-time traffic information uses the most up-to-date real-time information from traffic sensors and GPS data to create the most direct and efficient route to reach a destination. Real-time traffic information can be used to identify and avoid traffic jams, back-ups, and road works that may slow down traditional routes—ultimately resulting in a much faster and more efficient route.Leontiadis I. et al. [19] have discussed the effectiveness of an opportunistic traffic management system for vehicular networks is Limited. It relies on available capacity in the network to determine when and where vehicles should be allowed to transit. This means that the system is reactive—traffic congestion may still occur, and efficiency is determined by the amount of available capacity in the network.Further, these systems are vulnerable to inaccurate assumptions about the road network and the possibility of malicious interference with the transmissions. Doolan R. et al. [20] discussed the Vanet-enabled eco-friendly road characteristics-aware routing for vehicular traffic, a routing algorithm developed for determining an optimal route in congested road networks. The algorithm considers road characteristics such as length, width, number of lanes, turns, intersection types, average speed limit, traffic load, etc. It also considers the availability of green and sustainable transportation modes such as electric and hybrid vehicles. These factors calculate the fastest route with the least pollution impact. The goal of this routing algorithm is to reduce emissions, improve fuel efficiency, and minimize overall congestion. Nadeem T. et al. [21] have discussed Car-to-car communication, or vehicle-to-vehicle (V2V) communication, as a technology that wirelessly collects and shares data between vehicles on the road. This data is then used to facilitate communication between vehicles to share important information about traffic conditions, road hazards, and other timely information to help them operate safely on the roads. This technology can reduce traffic congestion and increase safety while driving.Yildirimoglu, M. et al. [22] has discussed the experienced travel time prediction for congested freeways is usually higher than the normal travel time. Congested freeways can experience extended travel times up to double the normal time.Wang Y., et al. [23] have discussed that Dynamic traffic prediction based on traffic flow mining is a method of predicting future traffic patterns by analyzing data mined from traffic flow sensors. This method utilizes machine learning algorithms, such as support vector machines, decision trees, neural networks, and deep learning, to create a model that can detect patterns in the data and make predictions about future traffic volumes and speeds. The predictive model can then be used to plan for changes in traffic flow, such as lane reductions and new road con-struction, and suggest optimal routes for drivers.Guardiola, I. G. et al. [24] has discussed the functional approach to monitoring and recognizing patterns of daily traffic profiles involves collecting a variety of data such as the amount of time spent on each website, the number of visitor interactions, the average speed of page loading, the amount of time spent on each page, and the type of pages visited. After collecting the data, it can then be analyzed to identify any patterns in the behavior of your users. For example, if you see a pattern of peak activity during certain times of the day, you can monitor the traffic and adjust your marketing strategy accordingly. You can also use this data to identify patterns that can be used to improve the user experience, such as speeding up page loading times or tailoring content to be more relevant to the user.Abdullah, S. M., et al. [25] has discussed the Soft GRU-Based Recurrent Neural Networks (RNNs) for Enhanced Congestion Prediction Using Deep Learning is a type of artificial neural network that uses deep learning techniques to predict traffic patterns and congested regions in urban areas. These soft GRU-based RNNs use transfer learning to learn complex patterns in the data quickly and accurately, allowing quick and accurate congestion prediction. These networks use various features such as historical traffic information, weather data, and geographical characteristics of urban areas to provide accurate predictions. The use of these networks for congestion prediction has helped make traffic control easier and more efficient.Logeshwaran J. et al. [26] have discussed a novel architecture ofanintelligent decision model for efficient resource allocation in 5G broadband communication networks is a system that integrates machine learning, optimization techniques, and real-time data analytics into an efficient resource allocation module. This system is designed to provide optimal utilization of network resources while meeting the demand of multiplex subscribers over a fixed time window. It incorporates predictive analysis tools to facilitate the dynamic allocation of resources based on historical data and user requirements. This model also enhances the scalability and reliability of 5G networks by allocating resources efficiently and without compromising the network's performance. Ultimately, this architecture is envisioned to provide optimal network performance while offering a flexible and efficient resource allocation with minimal user intervention.

Logeshwaran et al. [27] have discussed an energy-efficient resource allocation model for device-to-device communication in 5G wireless personal area networks that involves allocating radio resources to minimise energy consumption while achieving the desired data rate. Specifically, resource allocation strategies can include techniques such as frequency reuse, power control, and scheduling of transmissions to maximize channel utilization and minimize energy consumption. Other methods used are modulation sub-carrier allocation, antenna configuration, and Beamforming. Energy-efficient resource allocation techniques consider the available spectrum utilization of the channel, the interference between users, traffic demand, and the mobility of users, and use advanced algorithms such as game theory, optimization theory, and machine learning to distribute the resources effectively.Singh et al. [28] have discussed Cloud-Based License Plate Recognition for Smart City Using Deep Learning. This system uses deep learning technology to identify car license plates from images captured in smart city surveillance cameras. This system enables automated license plate recognition (ALPR), increasing efficiency in traffic violation enforcement, enhancing public safety, and tracking vehicles to enforce traffic and security regulations. The system is cloud-based, meaning that data is stored and processed in the cloud and therefore does not require any physical hardware setup.

Additionally, this system can identify license plates from video streams, providingfaster and more accurate results than still images. Furthermore, deep learning models such as convolutional neural networks (CNN) allow this system to accurately recognize license plates in various conditions, including weather changes, different times of day, changes in lighting, and more.Wenget al. [29] have discussed a Decomposition Dynamic Graph Convolutional Recurrent Network (DDGCRN) for traffic forecasting, a deep learning model designed to predict future traffic conditions in a given region. DDGCRN combines a graph convolutional network (GCN) and a recurrent neural network (RNN) to capture both the spatial and temporal features of the traffic. The GCN part of the DDGCRN encodes the traffic data from different sources, such as traffic sensors, historical data, and geographical information, into a vector representation. The RNN part of the DDGCRN then combines the vector representation and temporal features to forecast future traffic demand. For example, the RNN can consider holiday effects, seasonal changes, and other temporal factors.Djenouri et al. [30] discussed that Federated deep learning is a type of distributed machine learning technique that enables multiple parties to train a shared model securely without sharing their underlying data. It is increasingly used in edge-based applications in the smart city due to its data privacy, scalability, and cost-efficiency benefits. With federated deep learning, potentially sensitive data can stay within its original user's systems, allowing for improved shared insights and data-driven applications. By utilizing local edge devices, federated deep learning can enable applications ranging from waste management to advanced energy forecasting.

Walch, M. et al. [31] has discussed the Floating Car Data–Based Short-Term Travel Time Forecasting with Deep Recurrent Neural Networks. Incorporating Weather Data is a forecasting methodology that uses Floating Car Data (FCD) combined with deep recurrent neural networks and weather data to predict travel time for short-term trips. This approach combines data from the floating car (e.g., location) and local and global weather information to provide better prediction accuracy than traditional models that lack a weather factor. By incorporating weather data into the model, the system can make predictions that are more accurate than those of traditional methods. Furthermore, this approach can capture much finer route detailsthan other models, allowing it to provide more specific predictions for each vehicle.Maheswari, K. G., et al. [32] has discussed the Optimal cluster-based feature selection for intrusion detection systems in web and cloud computing environments using hybrid teacher learning optimization, and deep recurrent neural networks can occur through various methods. These methods include expert knowledge-based rules, feature selection algorithms, and random forest-based models. Expert knowledge-based rules utilize domain-specific knowledge, rules, and other manual information to identify the important features of the dataset. Feature selection algorithms use mathematical models to determine the optimal feature subset from the dataset. Random forest-based models use decision tree-like structures to identify important features with the strongest predictor ability. All these methods will help identify optimal features for intrusion detection systems and improve the performance of intrusion detection systems, especially in a dynamic and unpredictable environment.Rezaee et al. [33] discussed the IoMT-Assisted Medical Vehicle Routing Based on UAV-Borne Human Crowd Sensing and Deep Learning in Smart Cities is a type of intelligent optimization approach which combines Internet of Things (IoT) technology, UAV-borne human crowd sensing, and deep learning technology to help guide medical vehicles to their destinations more efficiently. It utilizes IoT sensors installed on medical vehicles to collect road data such as traffic conditions and congestion to dynamically generate optimized routes. It combines UAV-borne human crowd sensing to collect updated information about the real-time situation on roads or in areas of interest. Finally, it utilizes deep learning to analyze the collected data to generate intelligent recommendations and alerts that enable medical vehicles to make more accurate decisions on routes to maximize efficiency and reduce travel time. This approach is especially useful in smart cities where traffic conditions are constantly changing.Asha, A., et al. [34] has discussed the Optimized RNN-based performance prediction of IoT and WSN-oriented smart city application using an improved honey badger algorithm is an advanced method of predicting the performance of smart city applications using an enhanced version of the honey badger algorithm and a Recurrent Neural Network. This improved version of the honey badger algorithm combines reinforcement learning, dynamic programming, and evolutionary computation to increase the accuracy of the performance predictions. The Re-current Neural Network, part of the method, is used to handle the temporal elements of the prediction and increase the accuracy of the predictions.

Jain et al. [35] discussed the Research on artificial neural networks (ANNs) for smart cities towards Sustainable Development Goal (SDG)-11 has grown rapidly in recent years due to the potential for improved services and outcomes in urban areas. Research themes include the development of ANN architectures for predictive analytics and forecasting, the development of ANNs for efficient use of energy resources and water, and the development of ANNs for improved governance and decision-making in smart cities. Other research themes include the development of ANNs for anomaly detection and data-driven urban planning. Innovative research trends include the development of ANNs that incorporate predictive models, such as Bayesian Networks, to identify complex problems and solutions in urban areas. Additionally, research on the use of ANNs for social network analysis and data-driven urban speciation has been conducted. Research concerning the role of ANNs in urban energy management and intelligent transportation systems is also an important area of focus. Vasudha et al. [36] discussed Carriageway edge detection for unmarked urban roads using deep learning techniques is the application of advanced computer vision and image processing techniques to identify and detect the edges of the road to aid an autonomous vehicle in its navigation while driving in an urban environment. The deep learning approach uses convolutional neural networks (CNNs) to analyse images or videos of the road ahead to detect objects, features, and other important cues that can be used to determine the route the car should take. This approach relies on accurately identifying features such as lane markings, spots, and other distinct features of the road to accurately identify and navigate around them. Nazari et al. [37] discussed that Deep learning has become an essential technology for the future of smart cities. It can be used to develop algorithms to solve complex problems such as traffic optimization, energy management, water quality management, and public safety. Deep learning can improve urban governance, plan smart city infrastructures, and optimize resource utilization. It can also be used to create more efficient and cost-effective smart city services such as parking management and smart waste management. Additionally, deep learning can help increase the accuracy and reliability of surveillance data collected for public safety, and it can be used to develop advanced public services such as intelligent conversational agents. In short, deep learning can offer numerous advantages to improve people's overall quality of life in smart cities.

Abbas et al. [38] discussed that the Harris-Hawk-Optimization-Based Deep Recurrent Neural Network is designed to secure the Internet of Medical Things (IoMT) consisting of devices, such as implanted medical devices, connected to the Internet. The Harris-Hawk-Optimization-Based Deep Recurrent Neural Network is a bi-level optimization approach that combines deep recurrent neural networks with an optimization algorithm inspired by the behavior of the Harris Hawk. The optimization algorithm is based on a process known as the Hawk-Eye search, which allows the network to explore the input space more efficiently and accurately and to detect malicious activities more precisely. The network can identify suspicious missing data, detect suspiciously connected devices, identify malicious activity, and provide security alerts. The Harris-Hawk-Optimization-Based Deep Recurrent Neural Network also helps to protect against data leakage and unauthorized access. Xiao et al. [39] discussed the smart cities use a variety of predictive technologies to predict the availability of parking in a given area. This is done through the collection and analysis of data from sources such as sensors, traffic cameras, and vehicle tracking software. The predictive models developed by smart cities help city planners and decision-makers plan the best parking strategies for a given area, such as dynamic pricing, improved enforcement, or creating new parking spaces altogether. Predictive models also allow drivers to locate and reserve parking spots in real time easily.

Redhu et al. [40] discussed the Short-term traffic flow prediction based on optimized deep learning neural networks, which use deep learning algorithms to optimize traffic flow prediction accuracy. It involves building neural networks with multiple layers to learn from historical traffic data to predict future traffic. The prediction process is iterative, and the deep learning algorithm can adapt to changes in traffic flow and trends. This type of prediction helps traffic engineers manage congested roadways and advise drivers on the best route.Sereyet al. [41] discussed the Pattern recognition and deep learning technologies are enablers of Industry 4.0, or the fourth industrial revolution. They enable machines that can rapidly analyze data and make decisions previously limited to human brains, from sorting through large volumes of data to completing complex tasks. They can identify trends, predict outcomes, and generate insights with unprecedented speed and accuracy.Pattern recognition and deep learning can be used in various industries, such as computer vision, natural language processing, robotics, driverless cars, healthcare, and finance. By combining large amounts of data from multiple sources, these technologies allow for more intelligent solutions and discoveries previously impossible.

Based on the above comprehensive analysis, the following issues were identified. They are,

- Incomplete or biased data interpreting traffic volume and Variation in Traffic Congestion Levels across Different Periods
- Complexity in Differentiating Street Traffic Types and Data Security and Privacy Considerations
- Lack of Standardization of Systems and limited Accessibility of High-Quality Historical Data
- Difficulties in Modeling for Unforeseen Events and Geographical Constraints on System Performance
- Inaccuracy of Interpretation for Motorists on Pedestrians, or Cyclists and the limitations in Updating Emergency Services in Timely Manner

The novelty of this approach is that it uses a deep recurrent neural network for traffic pattern classification in smart cities. Compared to traditional methods, this allows for a more accurate and comprehensive classification of traffic patterns, as the deep learning approach can learn patterns from sequences of data to identify, classify, and predict traffic patterns. Furthermore, the model can adapt to changes in the road network and other factors that may impact traffic, allowing for more accurate predictions. Additionally, the use of recurrent neural networks is advantageous as it can capture both spatial and temporal data, allowing for a more comprehensive approach.

## 3. Proposed Method

In this section, the proposed deep learning-based traffic pattern classification is developed. This is a route planning system that, at predetermined intervals, makes suggestions for traveler routes based on the present traffic status. It is possible for clients and servers to engage with one another. Theroute selection processis managed by the route server (shown in Figure 1) and is impacted by recent and past data in the route database.

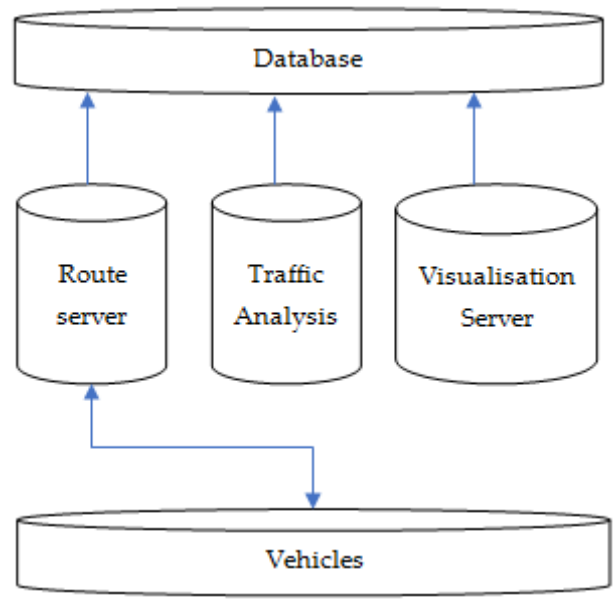

**Figure 1.** Proposed Framework.

The study can keep a close eye on the situation and make accurate forecasts about the current and future levels of traffic congestion by using the data gleaned from the server that performs traffic analysis and visualization. This allows them to monitor the situation in real-time. Figure 1 illustrates how client real-time feedback on traffic congestion circumstances improves the route database, which contributes to maintaining a real-time map of traffic in a city and accurate historical data on traffic behaviour. The predicted travel time for each path segment is no longer a constant based on the segment length and speed restriction; it is a value that varies dynamically during the day. This value was dependent on the segment length and speed restriction. This results in significantly better routes being supplied by this technology; however, this comes at the expense of an increased cost. The goal of DRNN is to use existing historical data on traffic records in a city to estimate trip times. Since these logs contain congestion data for each hour of the day for an entire city induction loop detectors for a year, the traffic analysis server needs to appropriately categorize them to be integrated into the route server without any delays. These logs contain data for an entire city's induction loop detectors. These logs contain data for the induction loop detectors throughout an entire city. This work operates on the traffic analysis aspect of the problem, and it proposes a DRNN that can simplify things by distilling massive amounts of traffic data from the past into a manageable set of representative daily patterns that can characterize the expected behaviour of traffic in the city at different times of the day.

*3.1. Traffic Max Computation*

The data on the volume of traffic is sent to us in the form of a time series vector, which is represented as $v \in R^{1 \times n}$. The number of time intervals and the number of road segments $m$ and $n$ are the two dimensions that can be observed in road networks. The spatial component refers to the number of road segments, while the temporal component refers to the number of road segments. Based on Markov chain theory, characterized by the use of transition matrices to express the probabilities associated with changing states, the notion of STM was developed. The STM is the notation used to express pairs of consecutive segments whenever there is a discussion regarding the probabilities of different speeds over a road network. In this study, the road network is modelled as a directed graph G = (V, E), where *V* is a collection of vertices representing intersections, and *E* is a set of edges representing road segments that connect two nearby intersections. The purpose is to investigate the relationship between the road network and the intersections that it contains. This modelling is being done to investigate the connection that exists between the road network and the intersections that it contains. The passage from edge ei to edge ej takes place at the precise instant t due to a shift in both space and time. The median speed is continuously monitored since it serves as an indicator of traffic flow. The calculated average speed along segment ej is denoted by sd, which stands for the end destination. The study considers the number of instances in which ei and ej are able to discretize their speed shifts from s0 to sd while continuing to perform their operations inside a given window of time t. Every result that was produced is a representation of the number of permutations that are conceivable between so and sd.

The counted number of speed changes is turned into a probability distribution over speed changes to calculate the likelihood of each transition. Both the resolution (sensitivity) of the speed change and the maximum speed that can be noticed play a part in selecting the dimensions of the values inserted into the matrix X. The maximum speed that can be observed is also a factor. The discretization value for speed in this investigation has been established at 5 kmph, and our maximum speed has been established at 100 kmph, producing the required matrix. The tests are conducted on stretches of road with speed limits varying from 50 to 80 kmph. A particular maximum speed figure is selected. This allows for greater accuracy in the results of the testing. The STM can be represented as Equation (1), where each value pij represents the chance that the vehicle speed was such and such at the beginning of the transition and sd at the end of the transition, at time t.

$$X = \begin{bmatrix} p_{11} & p_{12} & \cdots & p_{1n} \\ p_{21} & \ddots & \vdots & \vdots \\ \vdots & \vdots & \ddots & \vdots \\ p_{m1} & \cdots & \cdots & p_{mn} \end{bmatrix} \qquad (1)$$

The location of the pattern that was gathered is a key element of the data used to estimate the traffic situation.

*3.2. Deep Recurrent Neural Network*

A Long Short-Term Memory (LSTM) model was chosen as the deep learning model. When it comes to the process of learning data sequences, the long-term, short-term memory (LSTM) model performs extraordinarily well. The structure ensures that the knowledge learned in the past is mirrored in the information being learnt in the present. For this investigation, we use the model to arrange the flow-based learning data that documents traffic in the smart city in the appropriate temporal order.

*3.3. Multi-Layer LSTM*

LSTM networks are chosen for traffic classification due to their effectiveness in handling sequential data and variable-length inputs. Their memory retention and state preservation capabilities make them ideal for capturing long-term dependencies in traffic patterns. With applications in similar domains like speech and natural language processing, the multi-layer architecture allows for learning complex features. Its ability to handle sequential traffic data and capture long-term dependencies makes it suitable for accurate traffic classification in smart cities.

The design of an LSTM network is capable of receiving inputs of any length, and it may be implemented in many different ways to fulfil the requirements of a wide range of applications. LSTM networks are used in areas such as speech recognition and image recognition. An LSTM architecture with multiple layers is used for this research. The multi-layer LSTM model requires a specific number of successive packets for each flow when trained on a flow-based dataset. In the dataset based on flows, the initial packet of each flow is placed in the LSTM layer cell designated for that flow. The input for a new packet received at the input is taken from the output of the first LSTM cell. This occurs whenever a new packet is received at the input. The result of what happens in the first cell affects the method by which the functions of the second cell are carried out. The operation's output carried out by the first cell is further delivered as an input to the second layer of the LSTM (Table 1).

**Table 1.** Internal Operations (ms) of LSTM.

| Time Step | Input (Packet) | First LSTM Output | Second LSTM Input | Second LSTM Output |
|---|---|---|---|---|
| 1 | 0.2 | 0.1 | 0.1 | 0.08 |
| 2 | 0.3 | 0.15 | 0.15 | 0.11 |
| 3 | 0.5 | 0.2 | 0.2 | 0.14 |
| 4 | 0.1 | 0.12 | 0.12 | 0.09 |

The operation results carried out by the first cell are transferred to the second cell of the second LSTM layer. Performing this step is necessary to finish the circuit. Following this, the operation's output performed by the second cell in the third LSTM layer is transferred to the input of the third cell. Each LSTM cell contains an additional feature layer that is referred to as a cell state. The LSTM model can assess whether or not the weight value has been preserved. By utilizing the structures known as gates, the LSTM model can alter the state of its cells so that the information inside them may either be added to or removed from the system. A high level of persistence can be displayed by an LSTM since its gates can selectively accept or reject data. The remarkable overall performance of the LSTM model across the vast majority of tasks is partially attributable to the model's ability to fine-tune both the long-term memory and the final output.

*3.4. Model Tuning*

The values assigned to hyper-parameters define not only the network topology (for instance, the number of filters) but also the training method (e.g., type of optimizer). Because of the hyper-parameters that are chosen, the performance of a model is extremely subject to fluctuation. Specifically, we make use of grid search as the strategy for determining the appropriate hyperparameters for each deep learning model in accordance with the datasets. Grid search is an iterative search method. The term grid search can also be abbreviated as grid-search. When applied to a specific dataset, the grid search method does exhaustive testing on all the potential combinations of hyper-parameters to determine which hyper-parameter yields the best results and which hyper-parameter should be used moving forward. To maintain the model's accuracy, k-fold cross-validation is performed after the values that produce the best results for the model hyper-parameters have been determined. Before performing k-fold cross-validation on a dataset, the dataset must first be randomly divided into k sub-datasets that are roughly the same size as each other. One of the remaining k-subsets is set aside and assigned for use as validation data during the testing phase of the construction of the model. This happens while the other k-subsets are being utilized in training the model. We subjected the model to k iterations of cross-validation, with each of the k sub-datasets serving as validation data once throughout the process.

A single estimate can be derived by taking the average of each of the k different estimates provided. The practice of constantly selecting random sub-datasets to validate against is eliminated, which results in each observation only being used for validation exactly once. This constitutes an improvement over the previous method. Every observation is put to use, both for training and validation purposes.

The method utilized to ascertain whether or not a specific deep learning model possesses adequate hyper-parameters is illustrated in Figure 2. This can be accomplished by comparing the model to the data it used to train itself. The flow-based dataset is divided up into training samples and evaluation samples. The learning data are utilized to produce the training and validation sets, and then a k-fold cross-validation utilizing a grid search is performed. In addition to the hyper-parameter set provided, the model verification is performed based on the k-fold value. k is the number of times the value is multiplied by itself. After that, the optimal hyper-parameters are implemented into the training process of the model, and the test data are used to evaluate the accuracy of the model. In particular, when it comes to the multi-layer LSTM model, we make sure to consider hyper-parameters such as (1) output size, (2) kernel initialize, (3) recurrent initialize, (4) dropout rate, (5) output activation type, (6) optimization type, and (7) batch size. The dimensions of the output space are proportional to the output size in a direct and inverse connection. The kernel initialization is a method that enables the establishment of values for the kernel weight vector that will be applied to the inputs to perform the linear transformation. These values will be applied to the inputs to perform a linear transformation. The initial values for the regularization weight vector are assigned by the recurrent initialize. This responsibility lies with recurrent initialization.

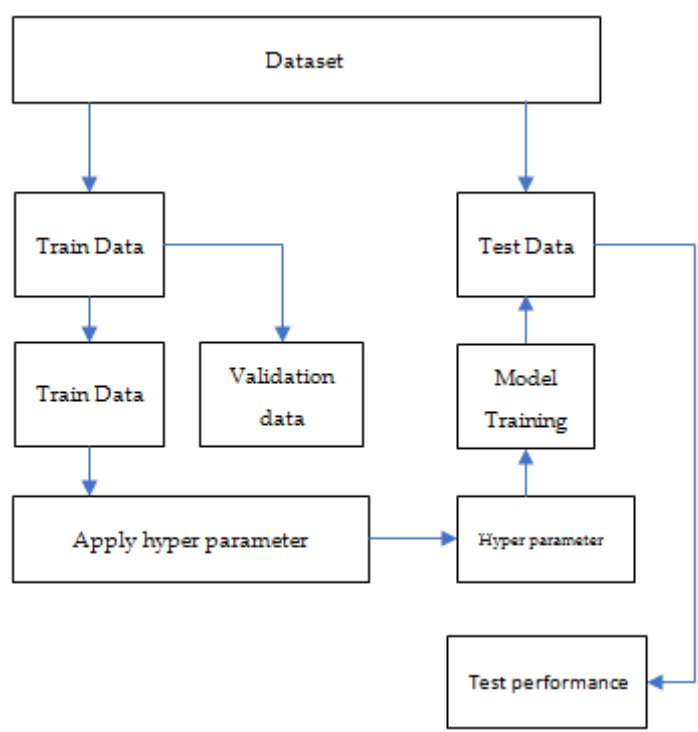

**Figure 2.** Data validation.

The dropout rate can be used to infer the proportion of hidden units that need to be eliminated before the recurrent state can be transformed linearly. This can be done by determining how many units have dropped out of the system. Find the proportion by deducting 100 from the total number of concealed units, then multiplying that result by 100. In all deep learning models, the definitions for output activation type, optimization type, and batch size are the same. For each size of the payload and the specified number of sequential packets in a flow, it would appear that the ideal parameter values for each hyper-parameter type in the multi-layer LSTM model are completely arbitrary. This is the case even though these values are supposed to be optimal. This is because of the optimum parameter settings for each hyperparameter type in the multi-layer LSTM model. There is a pattern to the values that work best for the parameters, and you may utilize this pattern to your advantage if you pay attention to it. To begin, the activation type used at the output is always SoftMax. This is the case,although the payload size or the packet count may change. Using a nonlinear logistic activation function is expected to result in the best outcomes when applying these models. The optimizer known as Adam typically results in the production of models of the highest possible quality. There is a connection between increasing the batch size and improving the model's performance.

## 4. Results and Discussion

Consider a map of a smart city with various roads and intersections displayed. The roads are color-coded to represent different traffic flow intensities, ranging from green for free-flowing traffic to red for heavy congestion. Alongside each road, numerical values indicate the current traffic volume, measured in vehicles per minute. For instance, a wide, multi-lane highway might be depicted in green with a traffic volume of 500 e-scooters per minute, suggesting smooth traffic flow. In contrast, a narrow city street could be shown in red with a traffic volume of 1500 e-scootersper minute, indicating severe congestion.

Datasets: Smart City traffic patterns are collected from the Kaggle repository [42]. For each of the day's eight-time intervals, the results of an attempt to estimate the present status of the traffic are displayed here. The values of the transitions labeled as crowded are at their greatest during rush hour, which was to be expected given the nature of the scenario because it is expected that more people will be using the transitions during rush hour. The percentage of the time that public transportation is congested is at its highest between peak hours. This points out that the congestion that started during the morning rush hour will continue during the following period and the following period after that one. This may also imply that the hours before and after rush hour see an increase in traffic congestion and that this issue continues to be a problem during the evening commute. This provides information regarding the overall accuracy of the model in addition to its precision, recall, and F1 scores for each class which are shown in Figures 3–6.

1. Accuracy Ratio: The accuracy improvement of the proposed method is expressed as a percentage ratio compared to the total number of transitions observed during the evaluation period. It shows how much the proposed method's predictions deviate from the ground truth regarding correct transitions.
2. Recall Ratio for the Free Flow Class: The recall ratio for the free flow class is calculated by dividing the number of correctly identified transitions in the free flow category by the total number of actual transitions that belong to the free flow class during the observation period. The result is a percentage ratio, indicating the proportion of correctly predicted free-flow transitions.
3. F1-score Ratio: The F1-score improvement is presented as a percentage ratio compared to the total number of transitions observed during the evaluation period. It combines both precision and recall metrics to provide a balanced evaluation of the proposed method's performance.
4. Computational and Communication Efficiency Ratio: The comparison of the computational and communication time of the proposed method against other methods is likely presented as a ratio. It indicates how much faster or slower the proposed method is in processing information compared to the total time taken by other methods.

The confusion matrix is a graphical representation of the efficiency of the classification system. The actual values are displayed in the rows of the matrix, while the projected class labels are shown in the columns of the matrix. The prediction accuracy is shown as a number in the matrix, and the contents of each cell in the matrix indicate the percentage of the available data for a class that was properly identified. When there are more than two classes that need to be differentiated, the level of precision can be calculated by dividing the total number of accurate classifications by the sum of the correct and incorrect classifications for all classes. This will give you a percentage that indicates how precise the classifications are. The F1 score can be calculated by finding the harmonic mean of the ratings for both precision and recall. A model's accuracy can be evaluated by taking the proportion of correct predictions, the fraction of incorrect predictions, and the sum of these three values across all classes and averaging them.

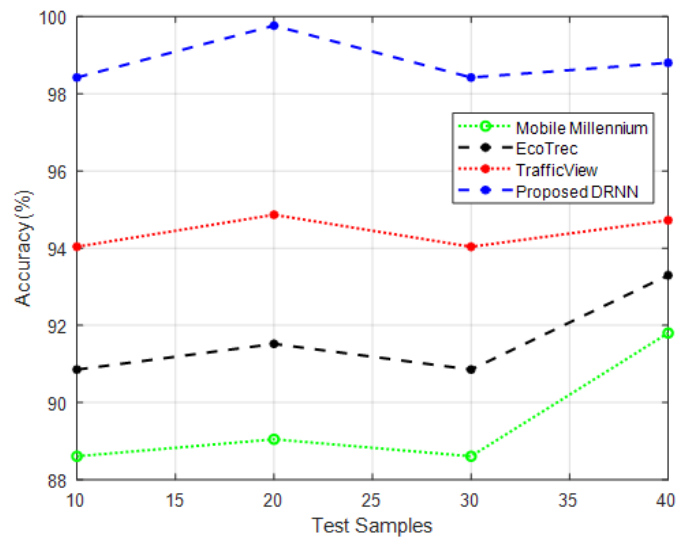

**Figure 3.** Computation of accuracy.

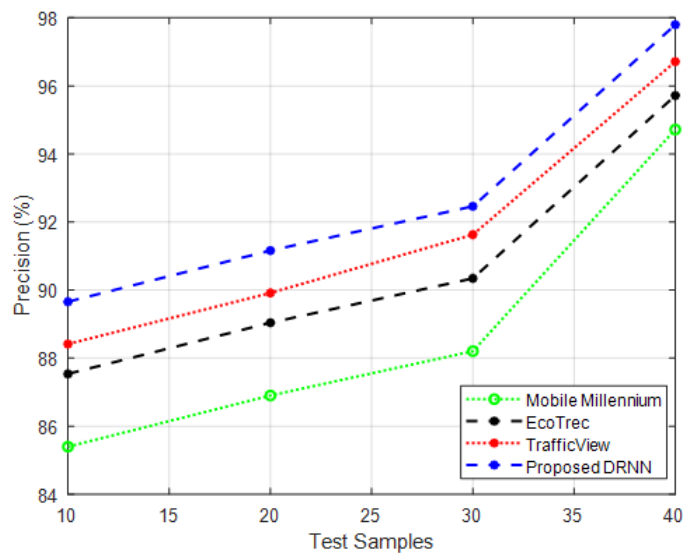

**Figure 4.** Computation of precision.

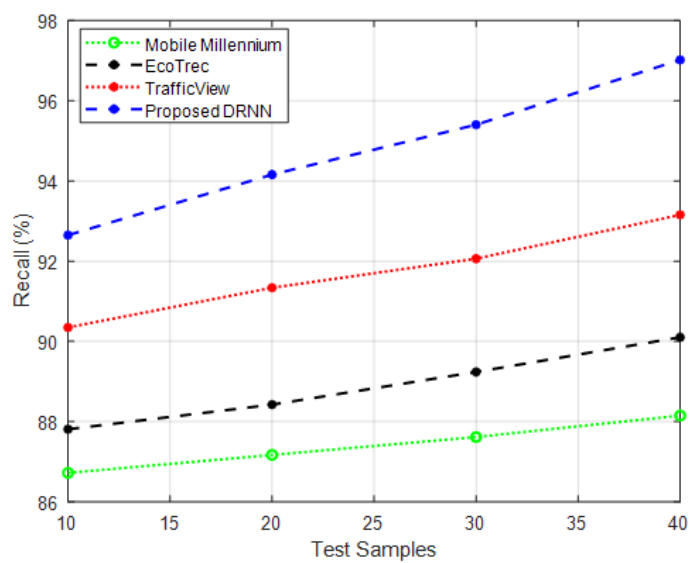

**Figure 5.** Computation of Recall.

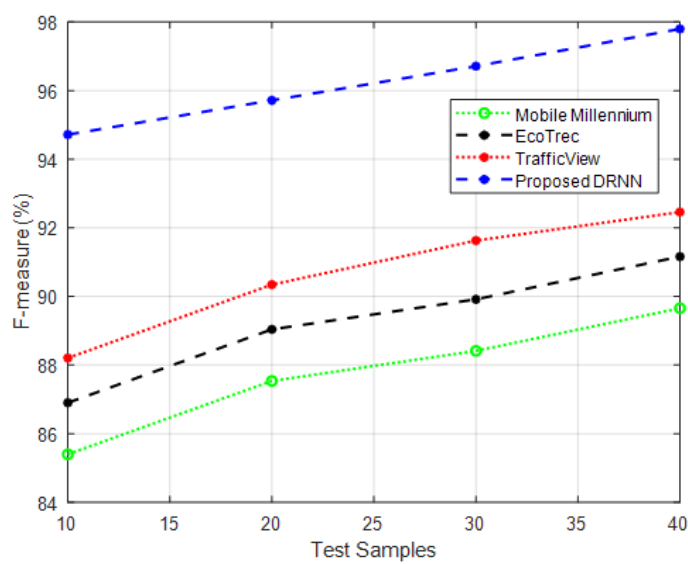

**Figure 6.** Computation of F1-score.

The comparison with the existing methods shows that the proposed method achieves a 3.5% of the percentage improvement in accuracy than the TrafficView model, 3.7% to the EcoTrec and 3.85% to Mobile Millennium. The completion of the validation process yielded an overall prediction accuracythat was 99% on average. The recall of the free flow class was only 90%, even though the precision predicts a perfect score, which is evidence that certain values were incorrectly assigned to the free flow class. The F1-score shows that the proposed method achieves a 3.2% of the percentage improvement in accuracy than the TrafficView model, 3.3% to the EcoTrec and 3.43% to Mobile Millennium. Tables2 and 3, show that the proposed method achieves reduced computational and communication time in processing the information than the other methods.

**Table 2.** Computational Time (ms).

| Sample | Proposed Method | TrafficView | EcoTrec | Mobile Millennium |
|---|---|---|---|---|
| 10 | 10.5 | 12.1 | 11.2 | 13.8 |
| 15 | 8.7 | 9.4 | 9.9 | 10.2 |
| 20 | 14.3 | 13.6 | 15.1 | 14.8 |
| 25 | 11.9 | 12.5 | 11.8 | 12.3 |
| 30 | 9.6 | 10.3 | 9.8 | 11.1 |
| 35 | 13.2 | 14.8 | 12.7 | 15.5 |
| 40 | 10.8 | 11.5 | 11.2 | 11.7 |

**Table 3.** Communication Time (ms).

| Sample | Proposed Method | TrafficView | EcoTrec | Mobile Millennium |
|---|---|---|---|---|
| 1 | 25 | 35 | 30 | 40 |
| 2 | 18 | 22 | 20 | 25 |
| 3 | 30 | 28 | 35 | 32 |
| 4 | 27 | 30 | 28 | 33 |
| 5 | 21 | 25 | 23 | 28 |
| 6 | 35 | 40 | 37 | 45 |
| 7 | 28 | 32 | 30 | 35 |

The proposed method, DRNN, for traffic classification in smart cities achieves superior accuracy, precision, recall, and F1-scoreresultscompared to existing state-of-the-art techniques. This success can be attributed to several key contributions of the DRNN approach: DRNN utilizes deep recurrent neural networks [43–45], which are well-suited for capturing the dynamic and sequential nature of traffic patterns. This allows the model to effectively analyze and understand the time-series traffic data, which is crucial for accurate traffic classification. The model employs convolutional and recurrent layers to extract essential features from the traffic data. The convolutional layer helps in spatial feature extraction, while the recurrent layer effectively captures temporal dependencies, resulting in a comprehensive representation of traffic patterns. Using a SoftMax layer enables the model to classify the traffic patterns effectively based on the extracted features. The SoftMax layer provides a probability distribution over different traffic classes, allowing the model to make more confident and accurate predictions. The proposed method introduces a heuristic approach to reduce the number of interpolation functions required to characterize common traffic scenarios. This optimization makes the modeling process more efficient and reduces the computational burden.

## 5. Conclusions

DRNN is the method of traffic classification in smart cities for this study. Deep recurrent neural networks are the basis of a novel approach that we detail here for the classification of traffic patterns. These networks are particularly well-suited to collecting the dynamic and sequential elements of traffic patterns, and that is the focus of our method. Those properties may be captured very effectively by using our method. To extract features from the aforementioned data on traffic patterns, the model that has been developed makes use of both a convolutional layer and a recurrent layer. In addition, a SoftMax layer is applied to classify the patterns that are recovered from the data. It provides the process to be followed to obtain realistic estimates of the amount of traffic congestion that exists on all of a city'sroutes throughout different times of the day and on each and every day of the year. Modeling takes a significant amount of time and resources; we came up with a heuristic to cut down on the number of interpolation functions necessary to adequately characterize common traffic scenarios. This was doneto make the modeling process more efficient. The proposed model achieves results that are superior in terms of accuracy, precision, recall, and F1 score compared to methods that are thought of as state-of-the-art techniques. In addition to this, we present a comprehensive analysis of the findings that the model may have for smart cities. The research indicates that the model that was proposed is capable of accurately classifying traffic patterns in smart cities to an accuracy level of up to 95%. After the proposed model has been applied to a dataset that contains actual traffic patterns to validate it, several other classification methods are usedto compare it to the results of the other models.

The proposed DRNN method for traffic classification in smart cities demonstrates impressive results but has some limitations. The model suffers from misclassification issues for the free flow class, indicating the need for parameter tuning or architectural adjustments. Additionally, data imbalance can bias the model's training and hinder performance in minority classes, necessitating data augmentation or weighted loss functions. The black-box nature of deep learning models presents challenges in interpreting the DRNN's decisions, calling for research in interpretability techniques. To ensure wider applicability, the method's generalization across diverse cities and real-time processing efficiency should be improved. Incorporating uncertainty estimation would enhance the model's ability to handle dynamic traffic conditions. Furthermore, integrating data from various sources can offer a more comprehensive understanding of urban traffic dynamics. Addressing these limitations and exploring new research directions in interpretability, generalization, real-time processing, and cross-modal integration will enhance the model's reliability and pave the way for more effective traffic classification in smart cities.

**Author Contributions:** Conceptualization, A.G.I. and K.J.; methodology, S.M.; investigation, S.A.; writing—original draft preparation, Y.N. and S.N.M.; writing—review and editing, Y.N. and S.N.M.; formal analysis, Y.N., S.A.; software, A.G.I.,K.J. and A.H.S.; visualization, S.M. and A.H.S.; supervision, validation, S.A; All authors have read and agreed to the published version of the manuscript.

**Funding:**This research is supported by Al-Kitab University, Kirkuk, Iraq.

**Institutional Review Board Statement:**Not applicable.

**Informed Consent Statement:**Not applicable.

**Data Availability Statement:** Not applicable.

**Conflicts of Interest:** The authors declare no conflict of interest.